\documentclass[sigconf]{acmart}

\usepackage{booktabs} % For formal tables

\setcopyright{rightsretained}
\acmDOI{10.475/123_4}

\acmISBN{123-4567-24-567/08/06}

\acmConference[PETRA'18]{ACM Woodstock conference}{June 26}{Corfu, Greece}
\acmYear{2018}
\copyrightyear{2018}

\acmArticle{4}
\acmPrice{15.00}

\begin{document}
\title{A Web Page Classifier Library Based on Random Image Content Analysis Using Deep Learning}

\author{Leonardo Espinosa Leal}
\orcid{1234-5678-9012}
\affiliation{%
  \institution{Arcada UAS}
  \streetaddress{Jan-Magnus Janssonin aukio 1}
  \city{Helsinki}
  \state{Finland}
  \postcode{00560}
}
\email{leonardo.espinosaleal@arcada.fi}

\author{Kaj-Mikael Bj\"{o}rk}
\affiliation{%
  \institution{Arcada UAS}
  \streetaddress{Jan-Magnus Janssonin aukio 1}
  \city{Helsinki}
  \state{Finland}
  \postcode{00560}
}
\email{kaj-mikael.bjork@arcada.fi}

\author{Amaury Lendasse}
\affiliation{%
  \institution{The University of Iowa}
  \city{Iowa City}
  \state{Iowa, US}
}

\author{Anton Akusok}
\affiliation{%
  \institution{Arcada UAS}
  \streetaddress{Jan-Magnus Janssonin aukio 1}
  \city{Helsinki}
  \state{Finland}
  \postcode{00560}
}

\renewcommand{\shortauthors}{L. Espinosa-Leal et al.}

\begin{abstract}
In this paper we present a methodology and the corresponding Python library\footnote{\url{https://bitbucket.org/akusok/wibc}} for the classification of webpages. Our method retrieves a fixed number of images from a given webpage, and based on them classifies the webpage into a set of established classes with a given probability. The library trains a random forest model build upon the features extracted from images by a pre-trained deep network. The implementation is tested by recognizing \emph{weapon} class webpages in a curated list of 3859 websites. The results show that the best method of classifying a webpage into the studies classes is to assign the class according to the maximum probability of any image belonging to this (weapon) class being above the threshold, across all the retrieved images. Further research explores the possibilities for the developed methodology to also apply in image classification for healthcare applications.
\end{abstract}




\maketitle

\section{INTRODUCTION}

Web classification is a hard but necessary task for different reasons: security, social conventions, parent control among others\cite{Vanetti2011,ali2017fuzzy}. It is also important for internet searchers because accurate classification can drive to better services. In the recent times a new application has emerged: Automating classification of healthcare information on the web~\cite{Cronin2015,SachioIOS2014}, were the main goal is to help users to avoid, filter or ranking web content during treatment of chronic diseases or improvement of habits for the treatment of internet addictions~\cite{GriffithTJSRs2001,MH15229}.

Traditional methods such as naive-Bayes or $k$-nearest neighbours have shown a good performance for the classification of web content~\cite{qi2009web}. However, the fast grown of the internet webpages mainly towards complex and interactive visual content makes the performance of these methods much poorer. In general, most of the techniques for the classification of webpages are based in the analysis of on-page features such as url name, textual content~\cite{milicka2013web} and tags~\cite{qi2009web} and more rarely visual content~\cite{deBoerSBH2011,burget2009web}. With the rise of deep learning as the method for image classification due to its accuracy, almost to human level, and the everyday more powerful computational resources in basic workstations have motivated the study of the visual features as a suitable source of information for classification of internet content.

In this work we classified webpages using the content images by means of a pre-trained neural network. The links of the images from a given webpage are obtained and then randomly shuffled. Then, a defined associated subset of all images is downloaded and then classified by our algorithm: A random forest model build upon the features of the pre-trained network. The whole process is carry out in memory. The final outcome is a list of probabilities linked to each image. These values are related to the pre-trained classes, which in our case is related with the weapon content. The presented method is highly accurate and can be easily extended to a large number of classes.

\section{Methodology}

The Website Image-Based Classification (WIBC) library written in Python 3.6 implements the novel functionality. It retrieves images from a given url and then calculates probabilities of belonging to a particular class for these images. Current study focuses on one relevant class: \emph{Weapons}, for which a large corpus of training and test data is available from the currently running CloSer project, but the methodology is extensible to any number of arbitrary classes. The deep learning back-end runs on an open source\footnote{Python package \url{https://pypi.python.org/pypi/mxnet},\\ accelerated version \url{https://pypi.python.org/pypi/mxnet-cu80mkl}} MXNet~\cite{MXNet} library. 

The whole url content of a webpage is accessed using the \emph{requests} python package. Links to all image content are extracted using the \emph{lassie} python package\footnote{http://lassie.readthedocs.io} from the url, and optionally from all possible sub-urls. Then image formatting function fetches images from these links and sets them in RGB format, dropping corrupted, inaccessible or tiny images. A search for image objects and their formatting continues iteratively until a required number of valid images is obtained. Such approach reduces url analysis latency, because not all links are followed. Valid images are returned as python image objects in a list, without a need for drive storage.

A parallel code implementation is written using a \emph{ProcessPool} job (from the \emph{Pebble} python package\footnote{https://pebble.readthedocs.io}). Distributed jobs are created incrementally as more image links are discovered, and the process pool terminates immediately after obtaining a necessary number of valid images. Parallel implementation does not support sub-url search, because it does not run with incremental image link discovery that greatly reduces a single url processing time.

In the next step, the associated features of each image are extracted using a deep network. Here we used a pre-trained\footnote{\url{https://github.com/dmlc/mxnet-model-gallery/blob/master/imagenet-21k-inception.md}} deep neural network formally known as Full ImageNet Network \emph{aka} Inception21k. This model is a pre-trained model on full imagenet dataset~\cite{5206848} with 14,197,087 images in 21,841 classes. The model is trained by only random crop and mirror augmentation. The network is based on Inception-BN network~\cite{pmlr-v37-ioffe15}. The final outcome is a \emph{numpy} array of dimension: 21,841$\times$number of images.

The 21,841 class predictions of the pre-trained model are used as inputs for our second stage model. It is trained on own set of images: a sanitized dump of random 120,342 images, and a set of 16,358 images from weapon websites (with a mix of weapon content and random content). A pure set of 793 manually sorted images of weapon content is used for testing. The class probability predictions are done by Random Forest Classifier from the Scikit-learn library~\cite{pedregosa2011scikit}. Therefore, in the last step, the matrix of extracted features is turned into weapon class probabilities by means of this model. 

The WIBC library works by instantiating a \emph{WIBC\_weapon} object, that loads a deep learning model and a random forest classifier into its memory. Then the \emph{run(url)} function fetches images from a website, and returns probabilities of them to belong to the \emph{Weapon} category. The \emph{WIBC\_weapon} settings include the maximum number of images to download per website, and whether to use GPU acceleration (with Nvidia CUDA 7.5 or 8.0).

\subsection{Methodology Under Development}

There are three directions of development for the nearest future. The first one extracts the upper-layer features from a deep network. These features provide universal and robust image description, that is independent of object classes and needs no re-training. The original network computes its classes from them with logistic regression. The proposed solution is to train a regularized linear model from these features towards our classes of interest. Then re-training a model on new dataset would involve only re-training the linear model, that is very fast.

The second approach is to store already observed images from each website (in some robust representation), and check all new images against them. Web images are highly repetitive across webpages. Knowing that an image was observed on a website of a particular class provides valuable classification information, may be faster than a deep network, and works with any hard-to-classify image content like \emph{Cults} website.

The third approach is to gather more training data from third-party image datasets. This involves running the WIBC library on large image datasets like Flickr1M (1 million images) to find probably relevant images to a particular category, and then manually filtering those images to extract the relevant ones. Another way to increase the training dataset are image rotation and mirroring.

Machine learning in health-care applications are a field that has been studied, but that also have great opportunities for further development. In Bj\"{o}rk et al. 2016~\cite{BjorkNAM2016}, the outline of the Huntington disease prediction project was presented, whereas the actual missing value imputation framework for more accurate prediction of the Huntington disease was presented by Akusok et al.~\cite{AkusokBMD2017}.

\section{Experimental Results}

\begin{figure}[t]
\centering
\includegraphics[width=0.4\textwidth]{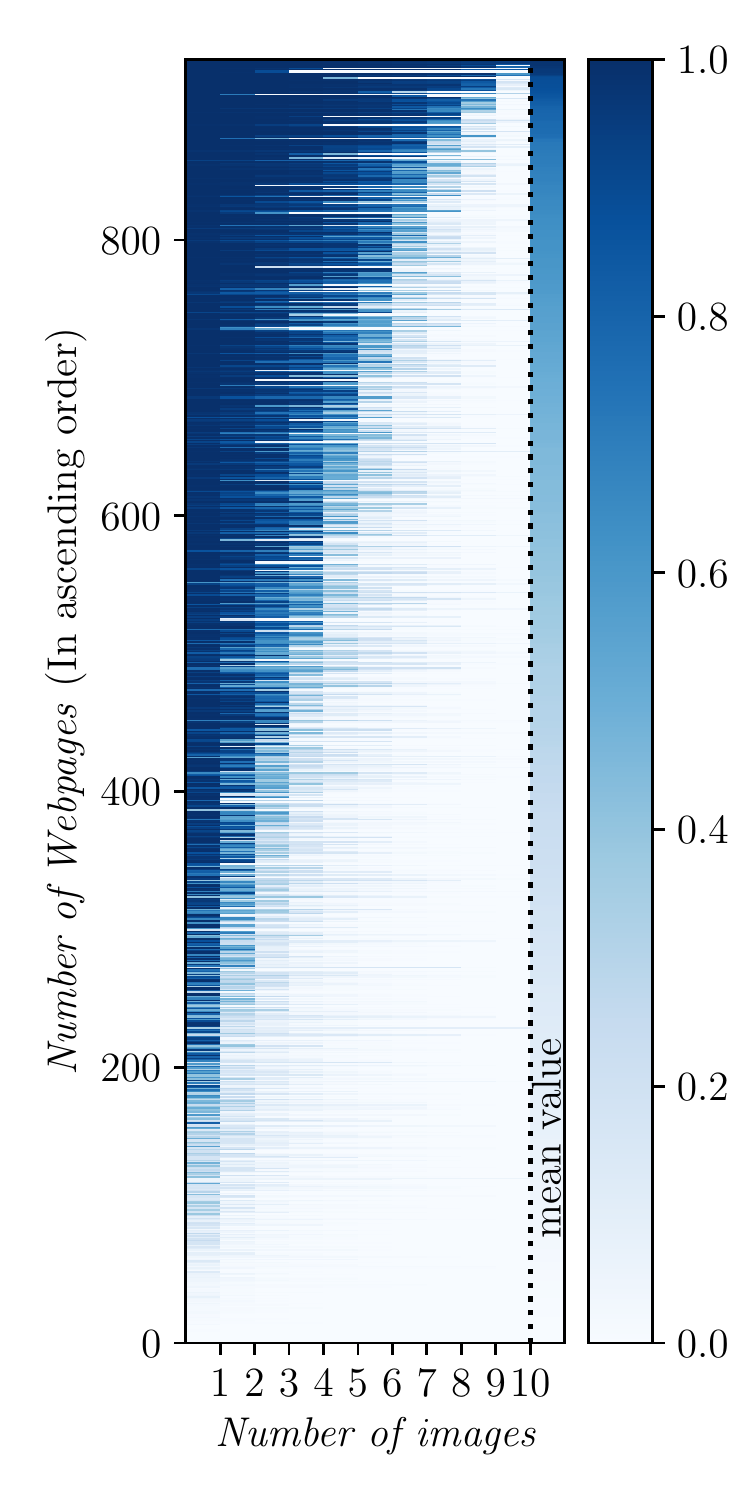}
\caption{\label{weapons}Result for a curated list of webpages with weapons content in ascending order.
We retrieved a maximum of ten images (when is possible) and assigned a probability of belonging to a weapons class. The mean value, assuming a Gaussian distribution of probabilities, is plotted on the right side of the main graph.}
\end{figure}

We tested our implementation on a curated list of webpages\footnote{F-Secure, private communication}. The list includes a collection of 1006 websites with content related to weapons, 1010 to alcohol, 715 to dating, 999 to shopping and 504 to random internet content (such as nasa.gov, google.com, etc). We did not have access, in average, to 9\% of the webpages mainly due to web crawling protection or non-existing domain. 

The results for the list of weapon related webpages is presented in the Fig.\ref{weapons}. The results were plotted in ascending order, depending of the mean value of the obtained probabilities of each webpage (see figure). We have analyzed a random subset of maximum ten images per webpage. The algorithm assigns a probability from 0 to 1 of belonging to a weapons class (from white to blue in the color scale). In average the result for a given webpage takes around 10 seconds, however this time can be tune depending of the user configuration (timeout, connection requests) and the internet speed connection. The download process can be done in parallel depending of the capabilities of the machine and for extracting the features of the images, a gpu option is included.

After collecting the results for all the webpages, the question to solve is what would be the criteria to decide if a webpage belongs to the weapons class or not? We have proposed two methods, first using the mean value of the probabilities assuming that for a given webpage these follow a Gaussian distribution, then we say if it belongs to the weapons class for a given threshold value. The Second method is classify the webpage if at least a $n$ number\footnote{$n$ can be 1, 2, 3, $\dots$ or 10., the maximum number of retrieved images.} of the obtained images belong to the weapons class for a given threshold.

\begin{figure}[t]
\centering
\includegraphics[width=0.475\textwidth]{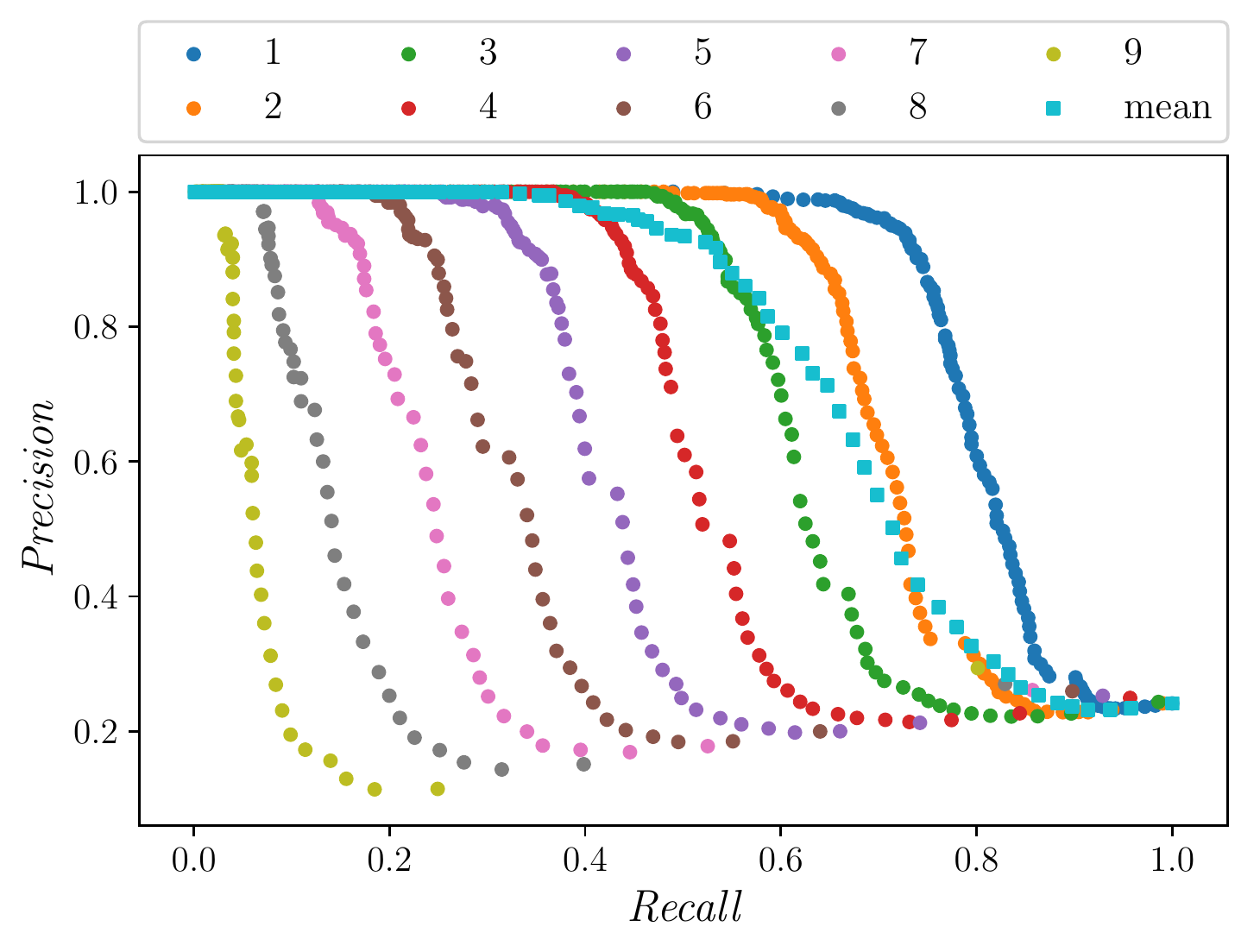}
\caption{\label{pre-rec} Precision-Recall curves for the different methods of measuring the presence of weapons in the studied websites. The first method is presented in magenta-squares and the second method for different values of $n$, from 1 to 9 is presented in colored circles.}
\end{figure}

\begin{figure}[t]
\centering
\includegraphics[width=0.475\textwidth]{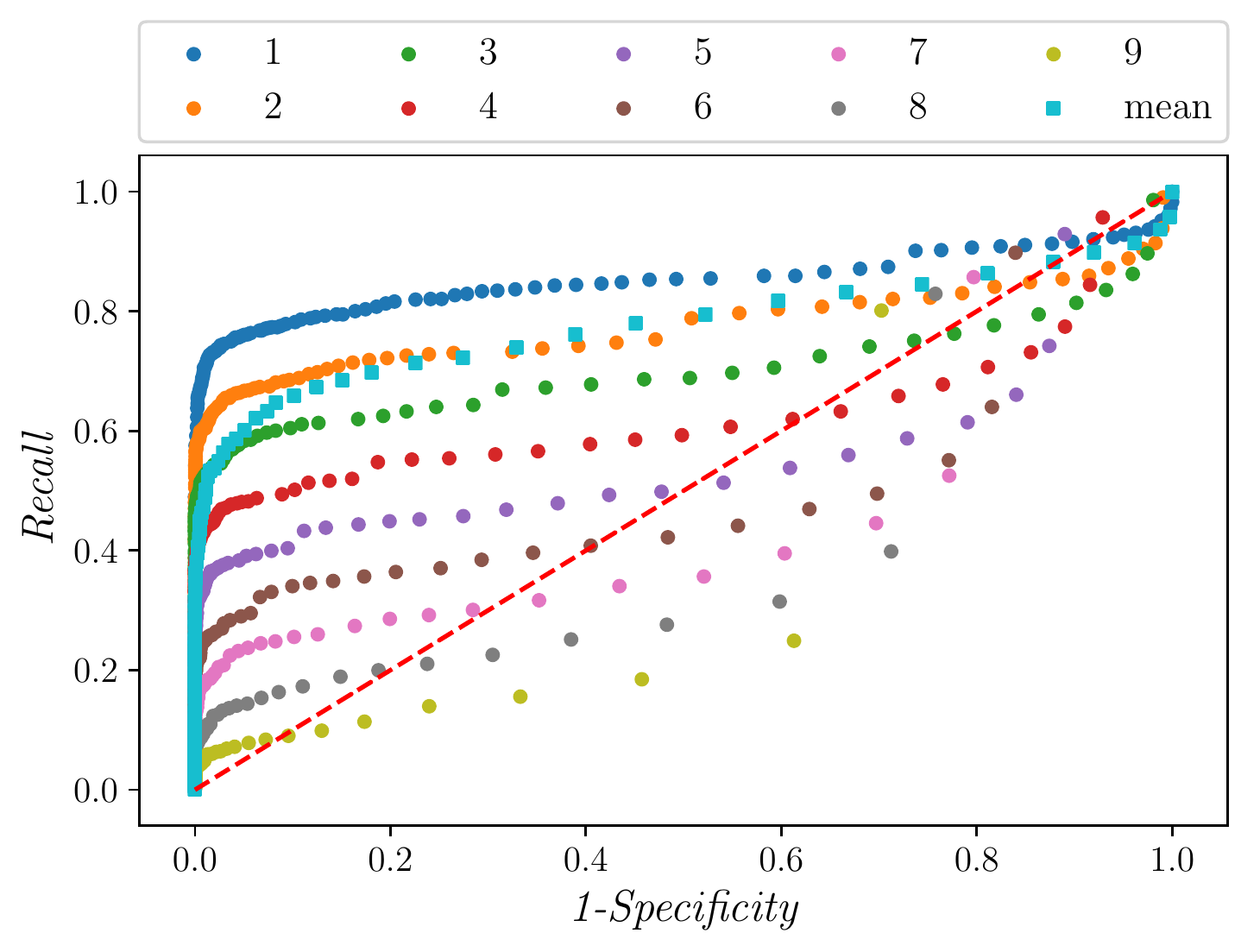}
\caption{\label{roc_curves} ROC curves for the different methods for measuring the presence of weapons in the studied websites. The first method is presented in magenta-squares and the second method for different values of $n$, from 1 to 9 is presented in colored circles.}
\end{figure}

In the Figs.\ref{pre-rec} and \ref{roc_curves} appear plotted the Precision-Recall and ROC curves for the two methods. The curves for the first method is plotted in magenta-squares meanwhile the for the second method, for each value of $n$, the result is plotted with colored circles. 

Following the plots, it is clear that the best method, with the best Precision-Recall or less false positive rate consists, is the second proposed method with $n$=1. The respective balance point (where Sensitivity = Specificity) is equal to 0.81 and it is obtained when the probability threshold is around 0.41. The second best result is obtained with the second method at $n$=2. The first method is ranked third.

\section{Conclusion and further research}

In the classification of webpages, the analysis of the content of images is the best criteria so far. We propose that it is possible to classify a webpage in the weapons class by a single randomly selected image from the webpage belonging to this class. The threshold for this probability can be balanced depending of the precision or less false positive rate criteria.

Two highlights of the proposed method are high accuracy in image content analysis, and an extreme simplicity of implementation. The most complex part of image feature extraction is performed by a state-of-the-art deep learning model pre-trained on millions of images, that currently can be obtained with a single line of code~\footnote{\url{https://keras.io/applications}}. This saves weeks of development time on model structure selection, training parameter tuning and the training itself. Random Forest is a robust classifier that learns in seconds from arbitrary input features. The simplicity of implementation allows for flexible applications of the proposed method to any problem domain, and for easy scalability.

Further research includes exploration of the proposed method for classification of images in the health-care sector. In general, Machine learning techniques are very valuable in the health-care domain (e.g. the example of missing value imputation in~\cite{BjorkNAM2016,AkusokBMD2017}). In this case the possible application areas may be different from the Huntington disease prediction. As discussed, there are lots of diagnostic images processed in the health-care system like brain tumor detection~\cite{termenon_brain_2016}. In addition presumptive health-care processes, like filtering the content for people who try to quit smoking (for instance) can be of extreme value.

\section{ACKNOWLEDGMENT}
This material is based on work jointly supported by the Tekes project \emph{Cloud-assisted Security Services} (CloSer). The authors would like to thank their collaborators from F-Secure for sharing of the scientific research data.

%\begin{table}[h]
%\label{table_example}
%\begin{center}
%\begin{tabular}{|c||c|}
%\hline
%One & Two\\
%\hline
%Three & Four\\
%\hline
%\end{tabular}
%\end{center}
%\caption{An Example of a Table}
%\end{table}

%\newpage

%\section*{APPENDIX}

%If it is necessary.

%\section*{ACKNOWLEDGMENT}

%If we have some.

\bibliographystyle{ACM-Reference-Format}
\bibliography{mybibliography}


\begin{thebibliography}{17}

%%% the Web address.

\ifx \showCODEN    \undefined \def \showCODEN     #1{\unskip}     \fi
\ifx \showDOI      \undefined \def \showDOI       #1{#1}\fi
\ifx \showISBNx    \undefined \def \showISBNx     #1{\unskip}     \fi
\ifx \showISBNxiii \undefined \def \showISBNxiii  #1{\unskip}     \fi
\ifx \showISSN     \undefined \def \showISSN      #1{\unskip}     \fi
\ifx \showLCCN     \undefined \def \showLCCN      #1{\unskip}     \fi
\ifx \shownote     \undefined \def \shownote      #1{#1}          \fi
\ifx \showarticletitle \undefined \def \showarticletitle #1{#1}   \fi
\ifx \showURL      \undefined \def \showURL       {\relax}        \fi
% invisible to TeX
\providecommand\bibfield[2]{#2}
\providecommand\bibinfo[2]{#2}
\providecommand\natexlab[1]{#1}
\providecommand\showeprint[2][]{arXiv:#2}

\bibitem[\protect\citeauthoryear{Akusok, Eirola, Bj\"{o}rk, Miche, Johnson, and
  Lendasse}{Akusok et~al\mbox{.}}{2017}]%
        {AkusokBMD2017}
\bibfield{author}{\bibinfo{person}{Anton Akusok}, \bibinfo{person}{Emil
  Eirola}, \bibinfo{person}{Kaj-Mikael Bj\"{o}rk}, \bibinfo{person}{Yoan
  Miche}, \bibinfo{person}{Hans Johnson}, {and} \bibinfo{person}{Amaury
  Lendasse}.} \bibinfo{year}{2017}\natexlab{}.
\newblock \showarticletitle{Brute-force Missing Data Extreme Learning Machine
  for Predicting Huntington's Disease}. In
  \bibinfo{booktitle}{\emph{Proceedings of the 10th International Conference on
  PErvasive Technologies Related to Assistive Environments}}
  \emph{(\bibinfo{series}{PETRA '17})}. \bibinfo{publisher}{ACM},
  \bibinfo{address}{New York, NY, USA}, \bibinfo{pages}{189--192}.
\newblock
\showISBNx{978-1-4503-5227-7}


\bibitem[\protect\citeauthoryear{Ali, Khan, Riaz, Kwak, Abuhmed, Park, and
  Kwak}{Ali et~al\mbox{.}}{2017}]%
        {ali2017fuzzy}
\bibfield{author}{\bibinfo{person}{Farman Ali}, \bibinfo{person}{Pervez Khan},
  \bibinfo{person}{Kashif Riaz}, \bibinfo{person}{Daehan Kwak},
  \bibinfo{person}{Tamer Abuhmed}, \bibinfo{person}{Daeyoung Park}, {and}
  \bibinfo{person}{Kyung~Sup Kwak}.} \bibinfo{year}{2017}\natexlab{}.
\newblock \showarticletitle{A Fuzzy Ontology and SVM--Based Web Content
  Classification System}.
\newblock \bibinfo{journal}{\emph{IEEE Access}}  \bibinfo{volume}{5}
  (\bibinfo{year}{2017}), \bibinfo{pages}{25781--25797}.
\newblock


\bibitem[\protect\citeauthoryear{Bj\"{o}rk, Eirola, Miche, and
  Lendasse}{Bj\"{o}rk et~al\mbox{.}}{2016}]%
        {BjorkNAM2016}
\bibfield{author}{\bibinfo{person}{Kaj-Mikael Bj\"{o}rk}, \bibinfo{person}{Emil
  Eirola}, \bibinfo{person}{Yoan Miche}, {and} \bibinfo{person}{Amaury
  Lendasse}.} \bibinfo{year}{2016}\natexlab{}.
\newblock \showarticletitle{A New Application of Machine Learning in Health
  Care}. In \bibinfo{booktitle}{\emph{Proceedings of the 9th ACM International
  Conference on PErvasive Technologies Related to Assistive Environments}}
  \emph{(\bibinfo{series}{PETRA '16})}. \bibinfo{publisher}{ACM},
  \bibinfo{address}{New York, NY, USA}, Article \bibinfo{articleno}{49},
  \bibinfo{numpages}{4}~pages.
\newblock
\showISBNx{978-1-4503-4337-4}


\bibitem[\protect\citeauthoryear{Burget and Rudolfova}{Burget and
  Rudolfova}{2009}]%
        {burget2009web}
\bibfield{author}{\bibinfo{person}{R. Burget} {and} \bibinfo{person}{I.
  Rudolfova}.} \bibinfo{year}{2009}\natexlab{}.
\newblock \showarticletitle{Web Page Element Classification Based on Visual
  Features}. In \bibinfo{booktitle}{\emph{2009 First Asian Conference on
  Intelligent Information and Database Systems}}. \bibinfo{pages}{67--72}.
\newblock
\urldef\tempurl%
\url{https://doi.org/10.1109/ACIIDS.2009.71}
\showDOI{\tempurl}


\bibitem[\protect\citeauthoryear{Chen, Li, Li, Lin, Wang, Wang, Xiao, Xu,
  Zhang, and Zhang}{Chen et~al\mbox{.}}{2015}]%
        {MXNet}
\bibfield{author}{\bibinfo{person}{Tianqi Chen}, \bibinfo{person}{Mu Li},
  \bibinfo{person}{Yutian Li}, \bibinfo{person}{Min Lin},
  \bibinfo{person}{Naiyan Wang}, \bibinfo{person}{Minjie Wang},
  \bibinfo{person}{Tianjun Xiao}, \bibinfo{person}{Bing Xu},
  \bibinfo{person}{Chiyuan Zhang}, {and} \bibinfo{person}{Zheng Zhang}.}
  \bibinfo{year}{2015}\natexlab{}.
\newblock \showarticletitle{MXNet: {A} Flexible and Efficient Machine Learning
  Library for Heterogeneous Distributed Systems}.
\newblock \bibinfo{journal}{\emph{CoRR}}  \bibinfo{volume}{abs/1512.01274}
  (\bibinfo{year}{2015}).
\newblock


\bibitem[\protect\citeauthoryear{Coughlin, Williams, and
  Hatzigeorgiou}{Coughlin et~al\mbox{.}}{2017}]%
        {MH15229}
\bibfield{author}{\bibinfo{person}{Steven~S. Coughlin},
  \bibinfo{person}{Lovoria~B. Williams}, {and} \bibinfo{person}{Christos
  Hatzigeorgiou}.} \bibinfo{year}{2017}\natexlab{}.
\newblock \showarticletitle{A systematic review of studies of web portals for
  patients with diabetes mellitus}.
\newblock \bibinfo{journal}{\emph{mHealth}} \bibinfo{volume}{3},
  \bibinfo{number}{6} (\bibinfo{year}{2017}).
\newblock
\showISSN{2306-9740}


\bibitem[\protect\citeauthoryear{Cronin, Fabbri, Denny, and Jackson}{Cronin
  et~al\mbox{.}}{2015}]%
        {Cronin2015}
\bibfield{author}{\bibinfo{person}{Robert~M. Cronin}, \bibinfo{person}{Daniel
  Fabbri}, \bibinfo{person}{Joshua~C. Denny}, {and}
  \bibinfo{person}{Gretchen~Purcell Jackson}.} \bibinfo{year}{2015}\natexlab{}.
\newblock \showarticletitle{Automated Classification of Consumer Health
  Information Needs in Patient Portal Messages}.
\newblock \bibinfo{journal}{\emph{AMIA Annu Symp Proc}}  \bibinfo{volume}{2015}
  (\bibinfo{date}{05 Nov} \bibinfo{year}{2015}), \bibinfo{pages}{1861--1870}.
\newblock
\showISSN{1942-597X}
\newblock
\shownote{2245873[PII].}


\bibitem[\protect\citeauthoryear{de~Boer, van Someren, and Lupascu}{de~Boer
  et~al\mbox{.}}{2011}]%
        {deBoerSBH2011}
\bibfield{author}{\bibinfo{person}{Viktor de Boer}, \bibinfo{person}{MaartenW.
  van Someren}, {and} \bibinfo{person}{Tiberiu Lupascu}.}
  \bibinfo{year}{2011}\natexlab{}.
\newblock \bibinfo{booktitle}{\emph{Web Page Classification Using Image
  Analysis Features}}.
\newblock \bibinfo{publisher}{Springer Berlin Heidelberg},
  \bibinfo{address}{Berlin, Heidelberg}, \bibinfo{pages}{272--285}.
\newblock


\bibitem[\protect\citeauthoryear{Deng, Dong, Socher, Li, Li, and Fei-Fei}{Deng
  et~al\mbox{.}}{2009}]%
        {5206848}
\bibfield{author}{\bibinfo{person}{J. Deng}, \bibinfo{person}{W. Dong},
  \bibinfo{person}{R. Socher}, \bibinfo{person}{L.~J. Li}, \bibinfo{person}{Kai
  Li}, {and} \bibinfo{person}{Li Fei-Fei}.} \bibinfo{year}{2009}\natexlab{}.
\newblock \showarticletitle{ImageNet: A large-scale hierarchical image
  database}. In \bibinfo{booktitle}{\emph{2009 IEEE Conference on Computer
  Vision and Pattern Recognition}}. \bibinfo{pages}{248--255}.
\newblock
\showISSN{1063-6919}


\bibitem[\protect\citeauthoryear{Griffiths}{Griffiths}{2001}]%
        {GriffithTJSRs2001}
\bibfield{author}{\bibinfo{person}{Mark Griffiths}.}
  \bibinfo{year}{2001}\natexlab{}.
\newblock \showarticletitle{Sex on the internet: Observations and implications
  for internet sex addiction}.
\newblock \bibinfo{journal}{\emph{The Journal of Sex Research}}
  \bibinfo{volume}{38}, \bibinfo{number}{4} (\bibinfo{year}{2001}),
  \bibinfo{pages}{333--342}.
\newblock


\bibitem[\protect\citeauthoryear{Hirokawa and Ishita}{Hirokawa and
  Ishita}{2014}]%
        {SachioIOS2014}
\bibfield{author}{\bibinfo{person}{Sachio Hirokawa} {and} \bibinfo{person}{Emi
  Ishita}.} \bibinfo{year}{2014}\natexlab{}.
\newblock \bibinfo{booktitle}{\emph{Non-topical classification of healthcare
  information on the web}}. \bibinfo{series}{Frontiers in Artificial
  Intelligence and Applications}, Vol.~\bibinfo{volume}{262}.
\newblock \bibinfo{publisher}{IOS Press}, \bibinfo{address}{Netherlands},
  \bibinfo{pages}{237--247}.
\newblock
\showISBNx{9781614994046}


\bibitem[\protect\citeauthoryear{Ioffe and Szegedy}{Ioffe and Szegedy}{2015}]%
        {pmlr-v37-ioffe15}
\bibfield{author}{\bibinfo{person}{Sergey Ioffe} {and}
  \bibinfo{person}{Christian Szegedy}.} \bibinfo{year}{2015}\natexlab{}.
\newblock \showarticletitle{Batch Normalization: Accelerating Deep Network
  Training by Reducing Internal Covariate Shift}. In
  \bibinfo{booktitle}{\emph{Proceedings of the 32nd International Conference on
  Machine Learning}} \emph{(\bibinfo{series}{Proceedings of Machine Learning
  Research})}, \bibfield{editor}{\bibinfo{person}{Francis Bach} {and}
  \bibinfo{person}{David Blei}} (Eds.), Vol.~\bibinfo{volume}{37}.
  \bibinfo{publisher}{PMLR}, \bibinfo{address}{Lille, France},
  \bibinfo{pages}{448--456}.
\newblock


\bibitem[\protect\citeauthoryear{Milicka and Burget}{Milicka and
  Burget}{2013}]%
        {milicka2013web}
\bibfield{author}{\bibinfo{person}{Martin Milicka} {and} \bibinfo{person}{Radek
  Burget}.} \bibinfo{year}{2013}\natexlab{}.
\newblock \showarticletitle{Web document description based on ontologies}. In
  \bibinfo{booktitle}{\emph{Informatics and Applications (ICIA), 2013 Second
  International Conference on}}. IEEE, \bibinfo{pages}{288--293}.
\newblock


\bibitem[\protect\citeauthoryear{Pedregosa, Varoquaux, Gramfort, Michel,
  Thirion, Grisel, Blondel, Prettenhofer, Weiss, Dubourg,
  et~al\mbox{.}}{Pedregosa et~al\mbox{.}}{2011}]%
        {pedregosa2011scikit}
\bibfield{author}{\bibinfo{person}{Fabian Pedregosa}, \bibinfo{person}{Ga{\"e}l
  Varoquaux}, \bibinfo{person}{Alexandre Gramfort}, \bibinfo{person}{Vincent
  Michel}, \bibinfo{person}{Bertrand Thirion}, \bibinfo{person}{Olivier
  Grisel}, \bibinfo{person}{Mathieu Blondel}, \bibinfo{person}{Peter
  Prettenhofer}, \bibinfo{person}{Ron Weiss}, \bibinfo{person}{Vincent
  Dubourg}, {et~al\mbox{.}}} \bibinfo{year}{2011}\natexlab{}.
\newblock \showarticletitle{Scikit-learn: Machine learning in Python}.
\newblock \bibinfo{journal}{\emph{Journal of Machine Learning Research}}
  \bibinfo{volume}{12}, \bibinfo{number}{Oct} (\bibinfo{year}{2011}),
  \bibinfo{pages}{2825--2830}.
\newblock


\bibitem[\protect\citeauthoryear{Qi and Davison}{Qi and Davison}{2009}]%
        {qi2009web}
\bibfield{author}{\bibinfo{person}{Xiaoguang Qi} {and}
  \bibinfo{person}{Brian~D. Davison}.} \bibinfo{year}{2009}\natexlab{}.
\newblock \showarticletitle{Web Page Classification: Features and Algorithms}.
\newblock \bibinfo{journal}{\emph{ACM Comput. Surv.}} \bibinfo{volume}{41},
  \bibinfo{number}{2}, Article \bibinfo{articleno}{12} (\bibinfo{date}{Feb.}
  \bibinfo{year}{2009}), \bibinfo{numpages}{31}~pages.
\newblock
\showISSN{0360-0300}


\bibitem[\protect\citeauthoryear{Termenon, Gra{\~n}a, Savio, Akusok, Miche,
  Bj{\"o}rk, and Lendasse}{Termenon et~al\mbox{.}}{2016}]%
        {termenon_brain_2016}
\bibfield{author}{\bibinfo{person}{Maite Termenon}, \bibinfo{person}{Manuel
  Gra{\~n}a}, \bibinfo{person}{Alexandre Savio}, \bibinfo{person}{Anton
  Akusok}, \bibinfo{person}{Yoan Miche}, \bibinfo{person}{Kaj-Mikael
  Bj{\"o}rk}, {and} \bibinfo{person}{Amaury Lendasse}.}
  \bibinfo{year}{2016}\natexlab{}.
\newblock \showarticletitle{Brain {{MRI}} Morphological Patterns Extraction
  Tool Based on {{Extreme Learning Machine}} and Majority Vote Classification}.
\newblock \bibinfo{journal}{\emph{Neurocomputing}}  \bibinfo{volume}{174, Part
  A} (\bibinfo{year}{2016}), \bibinfo{pages}{344 -- 351}.
\newblock
\showISSN{0925-2312}
\urldef\tempurl%
\url{https://doi.org/10.1016/j.neucom.2015.03.111}
\showDOI{\tempurl}


\bibitem[\protect\citeauthoryear{Vanetti, Binaghi, Carminati, Carullo, and
  Ferrari}{Vanetti et~al\mbox{.}}{2011}]%
        {Vanetti2011}
\bibfield{author}{\bibinfo{person}{Marco Vanetti}, \bibinfo{person}{Elisabetta
  Binaghi}, \bibinfo{person}{Barbara Carminati}, \bibinfo{person}{Moreno
  Carullo}, {and} \bibinfo{person}{Elena Ferrari}.}
  \bibinfo{year}{2011}\natexlab{}.
\newblock \bibinfo{booktitle}{\emph{Content-Based Filtering in On-Line Social
  Networks}}.
\newblock \bibinfo{publisher}{Springer Berlin Heidelberg},
  \bibinfo{address}{Berlin, Heidelberg}, \bibinfo{pages}{127--140}.
\newblock
\showISBNx{978-3-642-19896-0}


\end{thebibliography}

\end{document}